\documentclass{article}

\usepackage{array}
\usepackage[preprint]{neurips_2023}
\usepackage{multirow}

\usepackage[utf8]{inputenc} 
\usepackage[T1]{fontenc}    
\usepackage{hyperref}       
\usepackage{url}            
\usepackage{booktabs}       
\usepackage{amsfonts}       
\usepackage{nicefrac}       
\usepackage{microtype}      
\usepackage{xcolor}         

\usepackage{natbib} 
\setcitestyle{numbers}

\title{American Sign Language Video to Text Translation}

\author{%
    Parsheeta Roy \\
    Carnegie Mellon University\\
  \And
    Ji-Eun Han \\
    Carnegie Mellon University\\
    \AND
    Bhaavanaa Thumu \\
    Carnegie Mellon University \\
  \And
    Srishti Chouhan \\
    Carnegie Mellon University\\
}

\usepackage{graphicx}

\begin{document}

\maketitle

\begin{abstract}
 Sign language to text is a crucial technology that can break down communication barriers for individuals with hearing difficulties. We replicate and try to improve on a recently published study. We evaluate models using BLEU and rBLEU metrics to ensure translation quality. During our ablation study, we found that the model's performance is significantly influenced by optimizers, activation functions, and label smoothing. Further research aims to refine visual feature capturing, enhance decoder utilization, and integrate pre-trained decoders for better translation outcomes. Our source code  \footnote{https://github.com/jiSilverH/idlf23-aslt} is available to facilitate replication of our results and encourage future research.
\end{abstract}

\section{Introduction}
The task of Sign Language Translation (SLT) involves translating continuous sign language videos into spoken language sentences. It remains a challenging problem, requiring a precise understanding of signers' poses and generating textual transcriptions. 
Despite recent advances using deep neural networks trained on limited sign language datasets, the current state of automatic SLT is far from being solved.

To address this, research has now turned to larger and more complex datasets like How2Sign \cite{duarte2021how2sign}. The latest SLT baseline \cite{slt-how2sign-wicv2023} for this dataset, achieves a BLEU score of 8.03. However, it was found that relying solely on BLEU scores to select the best model checkpoint can be misleading. To address this issue, they proposed an alternative evaluation metric known as reduced BLEU (rBLEU), aiming for a more nuanced assessment of SLT system performance. rBLEU primarily considers semantically meaningful words thereby mitigating the risk of misleading model selection. 


In line with the methodologies outlined above, particularly the insights derived from the How2Sign dataset and the proposed utilization of rBLEU as an alternative evaluation metric, we intend to establish our SLT baseline. Leveraging the approaches and open-source resources detailed in this research, we aim to develop and fine-tune our ASL-to-English translation system.


\section{Literature Review} 

\subsection{Gloss-related Approaches in Sign Language Translation}
\textit{Sign glosses} correspond to spoken language words that represent the meanings of signs and, from a linguistic standpoint, appear as basic lexical elements. Glosses are typically used in linguistic research and analysis to provide a written approximation of how signs in a sign language correspond to words or concepts in spoken or written languages.


Previous research in SLT has explored the use of gloss supervision, where an intermediate textual representation is employed as an essential bridge between the input video sequence and the generated output text. This approach, called \textit{Gloss-based SLT} \cite{camgoz2020sign,  8578910, yin2020better}, has shown promise in enhancing the accuracy and efficiency of SLT systems by leveraging the power of glosses. However, the process of collecting and processing glosses is financially demanding and restricts data availability.

In contrast, our work focuses on a paradigm known as \textit{Gloss-free SLT}, which seeks to directly tackle the task of converting video into text without the reliance on any intermediate glosses. This gloss-free approach aims to overcome the limitations of gloss dependency, offering a more direct and potentially more scalable solution to SLT.

\subsection{Deep Learning-based Sign Language Translation}
Recent advancements have witnessed the integration of deep learning techniques, particularly those based on neural architectures, to enhance translation accuracy and efficiency. One common approach is the use of Recurrent Neural Networks (RNNs) with encoder-decoder architectures \cite{8578910, Fang_2017}, which have shown promise in SLT tasks. \cite{Fang_2017} uses a hierarchical bidirectional RNN for both word-level and continuous sign language translation. However, RNNs have inherent limitations in handling long-term dependencies, often resulting in sub-optimal translations. 
To address this challenge, the attention-based Transformer has emerged as a powerful solution in this field \cite{camgoz2020multichannel, yin2020better, chen2023twostream}. This architecture not only excels at modeling the temporal dependencies but also offers the advantage of parallelization during training, contributing to more efficient and effective sign language translation. \cite{yin2020better} employs two layers of transformers in contrast to the typical six layers used in standard spoken language translation.

\subsection{Tokenization of Sign Language Videos}
In the sign language translation system the input data consists of sign language videos, which are inherently dynamic and rich in visual information. To effectively capture the salient features from these videos, there are various state-of-the-art techniques such as 2D Convolutional Neural Networks (CNNs) \cite{8578910} and 3D ConvNets, particularly the Inflated 3D ConvNets (I3D) initially developed for action recognition \cite{albanie2021bsl1k}. Feature extraction can also be achieved through 2D pose estimation. In this context, \cite{li2020wordlevel} introduces an innovative approach with their Pose-based Temporal Graph Convolutional Networks (Pose-TGCN). This method simultaneously addresses both spatial and temporal aspects of human pose trajectories, thereby enhancing the effectiveness of pose-based techniques.

\subsection{Datasets}
The field of sign language research benefits from a variety of datasets, each offering unique contributions. The Purdue RVL-SLLL ASL Database \cite{1166987} is a notable resource, featuring 39 motion primitives with various hand shapes frequently used in American Sign Language (ASL). The Boston ASLLVD dataset is another valuable tool, comprising 2,742 words with a total of 9,794 examples.

In addition, RWTH-BOSTON-50 \cite{zahedi2005combination} includes 483 samples of 50 different glosses, RWTH-BOSTON-104 \cite{dreuw2008efficient} provides 200 continuous sentences encompassing 104 signs/words, and RWTH-BOSTON-400 \cite{dreuw-etal-2008-benchmark}, a sentence-level corpus, contains 843 sentences involving around 400 signs.

The How2Sign dataset \cite{duarte2021how2sign} stands out for its multimodality and multiview features, encompassing over 80 hours of continuous ASL videos and accompanying modalities like speech, English transcripts, and depth information. Lastly, the OpenASL dataset \cite{shi2022opendomain} offers 288 hours of ASL videos across multiple domains from over 200 signers, making it the largest publicly available ASL translation dataset to date. Each of these datasets plays a crucial role in advancing the study and understanding of sign language translation and interpretation.

\section{Model Description}
\subsection{Architecture}
The baseline model \ref{tab:model_architecture} is an asymmetric encoder-decoder comprising 6 encoder layers and 3 decoder layers, each with 4 attention heads, an embedding dimension of 256, a dropout of 0.3 and a feed-forward network of hidden size 1024. The adoption of a standard transformer encoder-decoder is driven by its ability to capture context and dependencies across input sequence, as well as its proficiency in sequence-to-sequence tasks.


\begin{table}[htbp]
  \centering
  \begin{tabular}{lp{0.6\linewidth}}
    \toprule
    \textbf{Component} & \textbf{Description} \\ [0.5ex] 
    \hline\hline
    \textbf{Encoder} &  \\
    \quad - \texttt{dropout\_module} & FairseqDropout \\
    \quad - \texttt{feat\_proj} & Linear: in\_features=1024, out\_features=256, bias=True \\
    \quad - \texttt{embed\_positions} & SinusoidalPositionalEmbedding \\
    \quad - \texttt{transformer\_layers} & List of TransformerEncoderLayer (0 to 5) \\
    \quad - \texttt{layer\_norm} & LayerNorm: (256,), eps=1e-05, elementwise\_affine=True \\ [1ex]
    \textbf{Decoder} &  \\
    \quad - \texttt{dropout\_module} & FairseqDropout \\
    \quad - \texttt{embed\_tokens} & Embedding: 7000 tokens, embedding\_dim=256, padding\_idx=1 \\
    \quad - \texttt{embed\_positions} & SinusoidalPositionalEmbedding \\
    \quad - \texttt{layernorm\_embedding} & LayerNorm: (256,), eps=1e-05, elementwise\_affine=True \\
    \quad - \texttt{layers} & ModuleList of TransformerDecoderLayerBase (0 to 2) \\
    \quad - \texttt{layer\_norm} & LayerNorm: (256,), eps=1e-05, elementwise\_affine=True \\
    \quad - \texttt{output\_projection} & Linear: in\_features=256, out\_features=7000, bias=False \\ [1ex]
    \textbf{TransformerEncoderLayer} &  \\
    \quad - \texttt{self\_attn} & MultiheadAttention: \\
    \quad \quad - \texttt{k\_proj} & Linear: in\_features=256, out\_features=256, bias=True \\
    \quad \quad - \texttt{v\_proj} & Linear: in\_features=256, out\_features=256, bias=True \\
    \quad \quad - \texttt{q\_proj} & Linear: in\_features=256, out\_features=256, bias=True \\
    \quad \quad - \texttt{out\_proj} & Linear: in\_features=256, out\_features=256, bias=True \\
    \quad - \texttt{self\_attn\_layer\_norm} & LayerNorm: (256,), eps=1e-05, elementwise\_affine=True \\
    \quad - \texttt{dropout\_module} & FairseqDropout \\
    \quad - \texttt{activation\_dropout\_module} & FairseqDropout \\
    \quad - \texttt{fc1} & Linear: in\_features=256, out\_features=1024, bias=True \\
    \quad - \texttt{fc2} & Linear: in\_features=1024, out\_features=256, bias=True \\
    \quad - \texttt{final\_layer\_norm} & LayerNorm: (256,), eps=1e-05, elementwise\_affine=True \\ [1ex]
    \textbf{TransformerDecoderLayerBase} &  \\
    \quad - \texttt{dropout\_module} & FairseqDropout \\
    \quad - \texttt{self\_attn} & MultiheadAttention: \\
    \quad \quad - \texttt{k\_proj} & Linear: in\_features=256, out\_features=256, bias=True \\
    \quad \quad - \texttt{v\_proj} & Linear: in\_features=256, out\_features=256, bias=True \\
    \quad \quad - \texttt{q\_proj} & Linear: in\_features=256, out\_features=256, bias=True \\
    \quad \quad - \texttt{out\_proj} & Linear: in\_features=256, out\_features=256, bias=True \\
    \quad - \texttt{activation\_dropout\_module} & FairseqDropout \\
    \quad - \texttt{self\_attn\_layer\_norm} & LayerNorm: (256,), eps=1e-05, elementwise\_affine=True \\
    \quad - \texttt{encoder\_attn} & MultiheadAttention: k\_proj, v\_proj, q\_proj, out\_proj \\
    \quad - \texttt{encoder\_attn\_layer\_norm} & LayerNorm: (256,), eps=1e-05, elementwise\_affine=True \\
    \quad - \texttt{fc1} & Linear: in\_features=256, out\_features=1024, bias=True \\
    \quad - \texttt{fc2} & Linear: in\_features=1024, out\_features=256, bias=True \\
    \quad - \texttt{final\_layer\_norm} & LayerNorm: (256,), eps=1e-05, elementwise\_affine=True \\ [1ex]
    \hline
  \end{tabular}
  \\ [1ex]
  \caption{Sign2Text Transformer Model Architecture}
  \label{tab:model_architecture}
\end{table}

\subsection{Methodology}
\begin{figure}[h!]
\centering
\includegraphics[width=7cm]{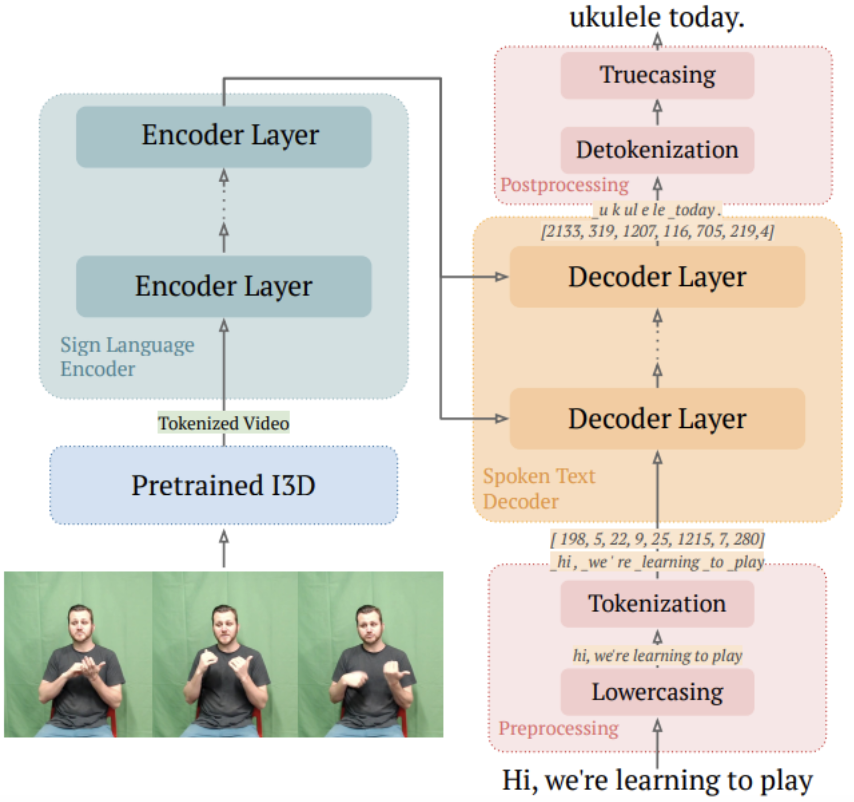}
\caption{The building blocks of the transformer model \cite{slt-how2sign-wicv2023}}
\label{fig:model}
\end{figure}
Inflated 3D ConvNet (I3D), is a deep learning architecture designed for video action recognition. 
I3D is constructed by inflating 2D convolutional filters into 3D, which are used to capture both spatial (height and width) and temporal features (frames over time) in videos. 
The 2D CNN model used as a basis is typically pre-trained on the ImageNet dataset.

As shown in Figure \ref{fig:model}, the input video stream undergoes tokenization using a pre-trained I3D feature extractor. These tokens are then fed to the encoder layers of the transformer. The transformer's decoder operates on lowercase and tokenized textual representations to generate the output text sequence.

\subsection{Evaluation Metrics}
\begin{figure}[h!]
\centering
\includegraphics[width=4cm]{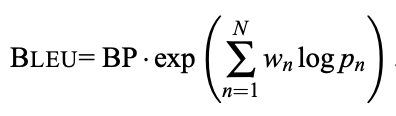}
\\
where the BP (Brevity Penalty) is\\
\includegraphics[width=4cm]{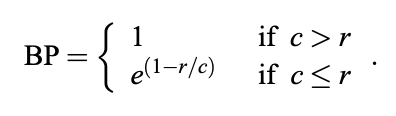}
\\
where c is the length of the candidate translation and r is the effective reference corpus length
\caption{Formula for BLEU score}
\label{fig:bleu_formula}
\end{figure}

Two distinct evaluation metrics are employed in the assessment. The BLEU score \cite{papineni-etal-2002-bleu} gauges the similarity between the predicted translation and the ground truth at the corpus level \ref{fig:bleu_formula}, while reduceBLEU (rBLEU) involves the exclusion of specific words from the reference and prediction prior to computing the BLEU score.

The challenging nature of SLT introduces a bias in the model's predictions toward the most statistically frequent patterns. These patterns can elevate BLEU scores without actually conveying meaningful translations when the prediction and reference share identical words, even if their meanings differ. 

Hence, the model evaluation employed rBLEU, an alternative metric to traditional BLEU, especially suitable for low-resource datasets containing frequent repetitive patterns. rBLEU offers a more accurate evaluation by mitigating the impact of these patterns on the score, providing a better reflection of the model’s actual performance compared to traditional BLEU.




\section{Loss Function}
The training process incorporates cross-entropy loss with label smoothing to enhance the model's generalization and expedite learning in the context of a sign-language translation network. 

Label smoothing changes the target vector by a small amount $\varepsilon$. Thus, instead of asking our model to predict 1 for the right class, we ask it to predict 1-$\varepsilon$ for the correct class and $\varepsilon$ for all the others. 

\begin{figure}[h!]
\centering
\includegraphics[width=7cm]{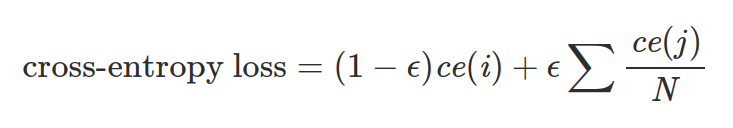}
\caption{Cross Entropy Formula with Label Smoothing}
\label{label_smoothing}
\end{figure}

In \ref{label_smoothing}, $ce(x)$ denotes the standard cross-entropy loss of $x$, $\varepsilon$ is a small positive number, $i$ is the correct class and $N$ is the number of classes.

By introducing soft targets derived from a weighted combination of hard targets and a uniform distribution across labels, this approach has proven effective in not only promoting generalization but also enhancing model calibration, a factor that significantly benefits beam-search algorithms \cite{müller2020does}. Label smoothing plays a pivotal role in encouraging tight clustering of representations for training examples belonging to the same class. 

\section{Baseline Implementation}
\subsection{Pre-processing}
I3D features are used to extract video representations directly from RGB frames and the I3D network resizes videos to $224 \times 224$. Color, scale, and horizontal flip augmentations are applied, and features are extracted from the 1024-dimensional activation before the pooling layer of the I3D backbone. 
As outputs of a trained network, these features do not require additional processing such as normalization.

Raw text is converted to lowercase to simplify the vocabulary and minimize the impact of irrelevant capitalization variations. A Sentencepiece tokenizer is then used to tokenize the lowercase text at the subword level with a vocabulary size of 7000 subwords. 

The selection of appropriate vocabulary size is a trade-off between enhancing the representation of rare words and producing shorter sequences. \cite{slt-how2sign-wicv2023} experimented with sizes of 1000, 4000 and 7000 sub-words and found that a larger dictionary improved results.

\subsection{Hyper-parameters}
In this paper, the following hyper-parameters were utilized for the experiments:
\begin{itemize}
  \item \textbf{Batch Size:} Maintained at \textbf{32} in order to reduce memory consumption for larger models

  \item \textbf{Learning Rate:} 
  \begin{itemize}
   \item \textbf{Initial Learning Rate:} Since the optimization of the learning rate is dependent on the number of parameters of the model, it was tuned together with other hyperparameters related to the architecture size.


   \item \textbf{Scheduler:}  Cosine with warm restarts as this has shown to perform better than alternatives. The resetting of the learning rate acts as a simulated restart of the learning process and is defined by the number of steps T.
  \end{itemize}


  \item \textbf{Number of Encoder-Decoder layers:} Experiments were conducted varying the number of encoder layers between 2, 3, 4, and 6, and the number of decoder layers between 2 and 3. This exploration likely aimed to assess how different layer counts affect the model's capacity to capture hierarchical features and contexts, balancing between model complexity and performance.



  \item \textbf{Number of Attention heads:} The number of attention heads were tested for configurations with either 4 or 8 heads, to analyze the trade-offs between computational complexity and the model's ability to learn intricate relationships. Results indicated that it was beneficial to use $4$ attention heads instead of $8$ under low-resource conditions.


  \item \textbf{Regularization:}  Adding dropout, weight decay, and label smoothing helps make the model
more robust to overfitting, considering it is difficult to perform data augmentation with video features.
  

  \item \textbf{Optimizer:} AdamW optimizer was used throughout the training due to its adaptive learning rates, momentum, and effectiveness in dealing with high-dimensional parameter spaces, facilitating faster convergence during training.
  
\end{itemize}

\subsection{Training}
After pre-processing, the model is trained for 108 epochs ($10^{5}$ steps) with a batch size of 32 and cross entropy loss with label smoothing of 0.1. 
The AdamW optimizer with a weight decay of 0.1 is used with a learning rate of 0.001 which is warmed up for the first 2000 updates. Subsequently, cosine decay is applied from $10^{-3}$ to $10^{-7}$ with warm restarts every 1.7 · $10^{4}$ steps.

During the text generation phase, the decoder predicts the next token by sampling from the probability distribution conditioned on previously generated tokens. The beam search algorithm (with a beam size of 5) is utilized to generate multiple candidate sequences, rather than selecting the highest probability prediction.

The output undergoes post-processing (detokenization and truecasing) to revert to the original capitalization. This ensures that the performance evaluation using the BLEU score is accurate.

We used the GCP T4 GPU and n1-highmem-4 for training, which took 11GB CUDA memory and ~6 hours for each ablation. We have also used the Nvidia RTX 3090 Ti GPU, which took ~3 hours to train per ablation and 22GB CUDA memory. It has been mentioned in \cite{slt-how2sign-wicv2023} that the training process took 3.5 hours on a NVIDIA GeForce RTX 2080 Ti GPU.

\subsection{Results}
\begin{table}[h!]
\centering
 \begin{tabular}{c c c c c c c} 
 \hline 
Partition & rBLEU & BLEU-1 & BLEU-2 & BLEU-3 & BLEU \\ 
 \hline\hline
 val & \textbf{2.79} & 35.2 & 20.62 & 13.25 & \textbf{8.89} \\ 
 test & \textbf{2.21} & 34.01 & 19.3 & 12.18 & \textbf{8.03}\\ 
 \hline
 \end{tabular}
 \\ [1ex]
 \caption{Reference paper scores on the How2Sign dataset}
\label{table:2}
\end{table}

\begin{table}[h!]
\centering
 \begin{tabular}{c c c c c c c} 
 \hline 
Partition & rBLEU & BLEU-1 & BLEU-2 & BLEU-3 & BLEU \\ 
 \hline\hline
 val & \textbf{2.75} & 35.883 & 20.956 & 13.345 & \textbf{8.924} \\ 
 test & \textbf{2.43} & 33.864 & 19.258 & 12.122 & \textbf{7.936}\\ 
 \hline
 \end{tabular}
 \\ [1ex]
 \caption{Baseline scores from our implementation on the How2Sign dataset}
\label{table:3}
\end{table}

Table \ref{table:2} presents the results reported in the paper by \cite{slt-how2sign-wicv2023} for the How2Sign dataset. In Table \ref{table:3}, we present the results obtained from our implementation, which involved training the model from scratch. 
The quantitative results demonstrate the machine translation metrics achieved by the implementation, serving as our baseline for further improvement.



\section{Experiments}
\subsection{Rationale}
The ablations conducted by \cite{slt-how2sign-wicv2023} are presented in Table \ref{table:4}. The authors identified a noteworthy degree of overfitting in larger models, likely stemming from the insufficient data provided relative to the necessary data volume required to fine-tune numerous parameters.

\begin{table}[h!]
\centering
\begin{tabular}{cccccccccccc}
\hline
ID & \begin{tabular}[c]{@{}c@{}}Encoder-\\ Decoder\\ Layers\end{tabular} & \begin{tabular}[c]{@{}c@{}}Embed\\ Dim\end{tabular} & \begin{tabular}[c]{@{}c@{}}FFN\\ Dim\end{tabular} & \begin{tabular}[c]{@{}c@{}}Attention\\ Heads\end{tabular} & Activation & LR    & Scheduler & rBLEU \\ \hline \hline
1  & 3-3 & 512 & 2048 & 8 & ReLU & 0.001 & inv sqrt & 0.98  \\ 
2  & 3-3 & 512 & 2048 & 8 & ReLU & 0.001 & cosine (T=17k) & 0.89  \\
3  & 3-3 & 256 & 1024 & 4 & ReLU & 0.001 & cosine (T=17k) & 1.14  \\
4  & 3-3 & 256 & 1024 & 4 & ReLU & 0.005 & cosine (T=17k) & 0.68  \\
\hline
5  & 2-2 & 256 & 1024 & 4 & ReLU & 0.001 & cosine (T=17k) & 1.32  \\
6  & 2-2 & 256 & 1024 & 4 & ReLU & 0.005 & cosine (T=17k) & 0.72  \\
7  & 2-2 & 256 & 512 & 4 & ReLU & 0.001 & cosine (T=17k) & 1.37  \\
\hline
8  & 4-2 & 256 & 1024 & 4 & ReLU & 0.001 & cosine (T=17k) & 1.14  \\
9  & 4-2 & 256 & 1024 & 4 & ReLU & 0.005 & cosine (T=17k) & 0.64  \\
\hline
10  & 6-3 & 512 & 2048 & 8 & ReLU & 0.001 & cosine (T=17k) & 0.87  \\ 
11  & 6-3 & 512 & 2048 & 8 & ReLU & 0.001 & cosine (T=22k) & 0.75  \\
12  & 6-3 & 256 & 1024 & 4 & ReLU & 0.001 & cosine (T=17k) & 0.93  \\
\hline
\end{tabular}
\\ [1ex]
\caption{Ablations conducted by \cite{slt-how2sign-wicv2023}}
\label{table:4}
\end{table}

However, after further experimentation, they concluded that a larger model, when combined with regularization techniques, outperforms a smaller model, as evidenced in Table \ref{table:5}.

\begin{table}[h!]
\centering
\begin{tabular}{cccccccccccc}
\hline
ID & \begin{tabular}[c]{@{}c@{}}Base\\ Model\end{tabular} & Dropout & \begin{tabular}[c]{@{}c@{}}Weight\\ Decay\end{tabular} & \begin{tabular}[c]{@{}c@{}}Label\\ Smoothing\end{tabular} & rBLEU \\ \hline \hline
13  & 1 & 0.1 & 0.001 & 0 & 0.98  \\ 
14  & 1 & 0.3 & 0.1 & 0.1 & 1.84  \\ 
15  & 7 & 0.2 & 0.01 & 0.1 & 1.21  \\ 
16  & 7 & 0.3 & 0.1 & 0.1 & 1.38  \\ 
17  & 12 & 0.2 & 0.01 & 0.1 & 1.17  \\ 
18  & 12 & 0.3 & 0.1 & 0.1 & \textbf{2.78}  \\ 
\hline
\end{tabular}
\\ [1ex]
 \caption{Validation Scores for Regularization experiments}
\label{table:5}
\end{table}

To further explore this theory and improve the baseline results, we conducted our ablations, drawing insights from the aforementioned experiments:
\begin{itemize}
  \item \textbf{Number of Layers:} Building on the success of \textbf{Model 18}, we conducted experiments with symmetric and asymmetric transformers, specifically focusing on configurations with 6 encoder layers and at least 3 decoder layers. 
  \item \textbf{Dimensions:} We retained the practise of using 4 and 8 attention heads with embedding dimensions of 256 and 512 respectively to prevent overfitting under low-resource conditions. 
  We also decided to test a 6-6 layer transformer with an FFN dimension of 512 since \textbf{Model 7} had the highest rBLEU score in the base ablations.
  \item \textbf{Activation:} The authors did not change their ReLU activation function so we decided to try GeLU instead since it has been reported to provide richer gradients for backpropagation \cite{Hendrycks2016GaussianEL} 
  \item \textbf{Learning Rate:} We maintained the initial learning rate at 0.001, as increasing it consistently degraded results. 
  We hypothesized that decreasing the learning rate would lead to slower convergence.
  \item \textbf{Scheduler:} Although,  inverse square root is a good default that produces decent results, a well-tuned cosine scheduler typically performs better in practice. 
  We can also observe that increasing the period for lr updates from $17k$ to $22k$ results in a dip in scores, likely caused by the learning rate decreasing too slowly, leading to suboptimal convergence in the later stages of training. 
  \item \textbf{Regularization:} As demonstrated in Table \ref{table:5}, increases in dropout, label smoothing, and weight decay significantly improved performance, particularly for larger models. 
  Consequently, we conducted ablations by incrementally increasing each parameter to analyze its impact on the overall score.
  \item \textbf{Batch Size and Optimizer:} We retained the batch size of 32 to ensure sufficient computational resources required for training large models and
  AdamW \cite{Loshchilov2017DecoupledWD} due to its generalization capabilities,  consistent performance and compatibility with learning rate schedulers.

\end{itemize}

\subsection{Ablations}

\begin{table}[h!]
\centering
\begin{tabular}{cccccccccccc}
\hline
ID & \begin{tabular}[c]{@{}c@{}}Encoder-\\ Decoder\\ Layers\end{tabular} & \begin{tabular}[c]{@{}c@{}}Embed-\\ FFN\\ (Attention)\end{tabular} & Activation & Dropout & \begin{tabular}[c]{@{}c@{}}Weight\\ Decay\end{tabular} & \begin{tabular}[c]{@{}c@{}}Label\\ Smoothing\end{tabular} & rBLEU \\ \hline \hline
19  & 6-3 & 512 - 2048 (8) & ReLU & 0.3 & 0.1 & 0.1 & 2.03 \\ 
20  & 6-3 & 512 - 2048 (8) & GeLU & 0.3 & 0.1 & 0.1 & 2.05 \\ 
21  & 6-3 & 512 - 2048 (8) & ReLU & 0.3 & 0.1 & 0.2 & 2.27 \\ 
22  & 6-3 & 512 - 2048 (8) & GeLU & 0.4 & 0.1 & 0.1 & 2.41 \\ 
23  & 6-3 & 512 - 2048 (8) & ReLU & 0.3 & 0.2 & 0.1 & 2.79 \\ 
24  & 6-3 & 512 - 2048 (8) & GeLU & 0.3 & 0.2 & 0.1 & \textbf{2.91} \\ 
25  & 6-3 & 512 - 2048 (8) & GeLU & 0.4 & 0.2 & 0.2 & 2.44 \\ 
\hline
26  & 6-6 & 256 - 512 (4) & ReLU & 0.3 & 0.1 & 0.1 & 2.33 \\ 
27  & 6-6 & 256 - 512 (4) & GeLU & 0.3 & 0.1 & 0.1 & 2.33 \\ 
28  & 6-6 & 256 - 512 (4) & ReLU & 0.4 & 0.1 & 0.1 & 2.40 \\  
29  & 6-6 & 256 - 512 (4) & ReLU & 0.3 & 0.1 & 0.2 & 2.54 \\ 
30  & 6-6 & 256 - 512 (4) & ReLU & 0.3 & 0.2 & 0.1 & 2.39 \\ 
\hline
31  & 6-6 & 256 - 1024 (4) & ReLU & 0.3 & 0.1 & 0.1 & 2.53 \\ 
32  & 6-6 & 256 - 1024 (4) & GeLU & 0.3 & 0.1 & 0.1 & 2.74 \\ 
33  & 6-6 & 256 - 1024 (4) & GeLU & 0.4 & 0.1 & 0.1 & 2.77 \\ 
34  & 6-6 & 256 - 1024 (4) & GeLU & 0.3 & 0.1 & 0.2 & \textbf{3.10} \\ 
35  & 6-6 & 256 - 1024 (4) & GeLU & 0.3 & 0.2 & 0.1 & \textbf{3.29} \\  
36  & 6-6 & 256 - 1024 (4) & GeLU & 0.3 & 0.2 & 0.2 & \textbf{2.92} \\ 
\hline
\end{tabular}
\\[1ex]
\caption{Validation scores across ablations}
\label{table:6}
\end{table}

The insights gleaned from our experimental findings, as detailed in Table \ref{table:6}, offer us a nuanced understanding of the model's behavior under varied configurations:
\begin{itemize}
    \item \textbf{Configuration 1 (Models 19 - 25):} 
      \begin{itemize}
       \item 
          Incorporating the same level of regularization as \cite{slt-how2sign-wicv2023} into the base \textbf{Model 10} results in a performance enhancement but falls short of matching the baseline results. This outcome is expected, considering the embedding and feedforward network dimensions are doubled.
        \item 
           The introduction of GeLU activation and the incremental adjustment of label smoothing and dropout values contributes to some performance improvement.
        \item 
          Notably, the most substantial performance improvement is observed when augmenting the weight decay. This indicates that the optimization process becomes complex for larger models and the increased weight decay results in a more stable and generalizable solution.
          \end{itemize}

    \item \textbf{Configuration 2 (Models 26 - 30):} 
      \begin{itemize}
       \item 
          Tripling the number of encoder-decoder layers of \textbf{Model 16} results in a significant improvement, likely due to the increased depth which enables the model to capture more complex relationships within the data.
        \item 
           However, introducing GeLU activation or increasing dropout or weight decay values result in little improvement, suggesting that the model adaptation may have saturated in this specific context.
        \item 
          The augmentation of label smoothing led to an improvement in model performance, which implies that label smoothing introduced an optimal level of regularization, considering the complexity of the model. However, the smaller FFN dimensions still seemed to limit the model's capacity to fully capitalize on this regularization, resulting in performance gains that could not surpass the baseline.
          \end{itemize}

    \item \textbf{Configuration 3 (Models 31 - 36):} 
    \begin{itemize}
    \item 
      Increasing the decoder layers of the best baseline \textbf{Model 18} inherently induces overfitting, which is exacerbated in the absence of supplementary regularization measures. 
    \item 
      Interestingly, the introduction of the GeLU activation function aligns the performance metrics with those of the baseline model.
    \item 
        The absence of a significant improvement when increasing the dropout from 0.3 to 0.4 suggests that the initial dropout rate may already be effectively regularizing the model, and further adjustments might not be conducive to improved performance in this scenario.
    \item
      Notably, increasing the values of label smoothing and weight decay results in a substantial improvement over the initial high score. This underscores the pivotal role of thoughtful regularization techniques in mitigating overfitting tendencies and optimizing model outcomes.
      \end{itemize}
      
\end{itemize}

\section{Results}

After an extensive sweep of hyperparameters and configurations, we achieved significant improvements which can be observed in Table \ref{table:7}.

\begin{table}[h!]
\centering
 \begin{tabular}{c c c c c c c c} 
 \hline 
Model ID & Partition & rBLEU & BLEU-1 & BLEU-2 & BLEU-3 & BLEU \\  
 \hline\hline
 \multirow{2}{*}{35} &
 val & \textbf{3.29} & 35.25 & 21.03 & 13.76 & \textbf{9.39} \\ &
 test & \textbf{2.56} & 33.20 & 19.12 & 12.10 & \textbf{7.95}\\  
 \multirow{2}{*}{36} &
 val & \textbf{2.92} & 35.85 & 21.50 & 14.04 & \textbf{9.59} \\ &
 test & \textbf{2.45} & 34.56 & 19.99 & 12.76 & \textbf{8.46}\\  
 \hline
 \end{tabular}
 \\ [1ex]
 \caption{Best validation and test scores on the How2Sign dataset}
\label{table:7}
\end{table}
Choosing more decoder layers, coupled with a smoother activation function and regularization methods was key to achieving optimal performance, resulting in a 0.51 rBLEU improvement. 
Furthermore, our results highlight the importance of finding the optimal point where regularization techniques help but do not hinder
the performance of the model. 


\section{Discussion}
In the original study \cite{slt-how2sign-wicv2023}, the authors initially established the importance of text preprocessing, larger vocabulary size and model architecture parameters in enhancing performance, demonstrating notable improvements through careful selection and experimentation. They identified overfitting in larger models due to insufficient data and addressed it by introducing optimal regularization techniques. Qualitative results indicated the model's ability to provide detailed translations for complex phrases but also highlighted challenges as high BLEU scores could be misleading due to frequent phrases.

We extended this research by retraining the baseline model and conducting additional ablations, while drawing insights from prior experiments. A comprehensive exploration of model configurations revealed nuanced relationships between various parameters and performance metrics. Key observations include the impact of regularization, layer dimensions, activation functions, and the interplay between different parameters.
In Configuration 1, increasing regularization demonstrated a notable improvement, with weight decay playing a crucial role. Configuration 2 highlighted the significance of layer width and depth, while Configuration 3 emphasized the delicate balance required for adjusting dropout, label smoothing, and weight decay to optimize model performance.
Furthermore, a strong correlation was observed between higher rBLEU scores and the model's efficacy in capturing semantic meaning from video content; highlighting the significance of performance metrics in evaluating the capabilities of a language model.




\section{Conclusion}
We explored \textit{sign language video-to-text translation}, which is a challenging task that aims to convert American Sign Language gestures into textual format. We successfully replicated the original implementation of \cite{slt-how2sign-wicv2023} and presented evaluation results using BLEU and rBLEU. We also conducted an extensive ablation study by systematically changing various hyperparameters which highlighted  the importance of careful parameter selection (model architecture, activation) and regularization techniques (dropout, weight decay and label smoothing) in mitigating overfitting and optimizing performance. Despite ongoing challenges, advancements in research and development hold promise for the future of sign-to-text translation, paving the way for more accurate and robust systems.


\newpage

\bibliographystyle{plainnat}
\bibliography{ref}

\begin{thebibliography}{19}
\providecommand{\natexlab}[1]{#1}
\providecommand{\url}[1]{\texttt{#1}}
\expandafter\ifx\csname urlstyle\endcsname\relax
  \providecommand{\doi}[1]{doi: #1}\else
  \providecommand{\doi}{doi: \begingroup \urlstyle{rm}\Url}\fi

\bibitem[Albanie et~al.(2021)Albanie, Varol, Momeni, Afouras, Chung, Fox, and Zisserman]{albanie2021bsl1k}
Samuel Albanie, Gül Varol, Liliane Momeni, Triantafyllos Afouras, Joon~Son Chung, Neil Fox, and Andrew Zisserman.
\newblock Bsl-1k: Scaling up co-articulated sign language recognition using mouthing cues, 2021.

\bibitem[Camgoz et~al.(2018)Camgoz, Hadfield, Koller, Ney, and Bowden]{8578910}
Necati~Cihan Camgoz, Simon Hadfield, Oscar Koller, Hermann Ney, and Richard Bowden.
\newblock Neural sign language translation.
\newblock In \emph{2018 IEEE/CVF Conference on Computer Vision and Pattern Recognition}, pages 7784--7793, 2018.
\newblock \doi{10.1109/CVPR.2018.00812}.

\bibitem[Camgoz et~al.(2020{\natexlab{a}})Camgoz, Koller, Hadfield, and Bowden]{camgoz2020multichannel}
Necati~Cihan Camgoz, Oscar Koller, Simon Hadfield, and Richard Bowden.
\newblock Multi-channel transformers for multi-articulatory sign language translation, 2020{\natexlab{a}}.

\bibitem[Camgoz et~al.(2020{\natexlab{b}})Camgoz, Koller, Hadfield, and Bowden]{camgoz2020sign}
Necati~Cihan Camgoz, Oscar Koller, Simon Hadfield, and Richard Bowden.
\newblock Sign language transformers: Joint end-to-end sign language recognition and translation, 2020{\natexlab{b}}.

\bibitem[Chen et~al.(2023)Chen, Zuo, Wei, Wu, Liu, and Mak]{chen2023twostream}
Yutong Chen, Ronglai Zuo, Fangyun Wei, Yu~Wu, Shujie Liu, and Brian Mak.
\newblock Two-stream network for sign language recognition and translation, 2023.

\bibitem[Dreuw et~al.(2008{\natexlab{a}})Dreuw, Forster, Deselaers, and Ney]{dreuw2008efficient}
Philippe Dreuw, Jens Forster, Thomas Deselaers, and Hermann Ney.
\newblock Efficient approximations to model-based joint tracking and recognition of continuous sign language.
\newblock In \emph{2008 8th IEEE International Conference on Automatic Face \& Gesture Recognition}, pages 1--6. IEEE, 2008{\natexlab{a}}.

\bibitem[Dreuw et~al.(2008{\natexlab{b}})Dreuw, Neidle, Athitsos, Sclaroff, and Ney]{dreuw-etal-2008-benchmark}
Philippe Dreuw, Carol Neidle, Vassilis Athitsos, Stan Sclaroff, and Hermann Ney.
\newblock Benchmark databases for video-based automatic sign language recognition.
\newblock In Nicoletta Calzolari, Khalid Choukri, Bente Maegaard, Joseph Mariani, Jan Odijk, Stelios Piperidis, and Daniel Tapias, editors, \emph{Proceedings of the Sixth International Conference on Language Resources and Evaluation ({LREC}'08)}, Marrakech, Morocco, May 2008{\natexlab{b}}. European Language Resources Association (ELRA).
\newblock URL \url{http://www.lrec-conf.org/proceedings/lrec2008/pdf/287_paper.pdf}.

\bibitem[Duarte et~al.(2021)Duarte, Palaskar, Ventura, Ghadiyaram, DeHaan, Metze, Torres, and i~Nieto]{duarte2021how2sign}
Amanda Duarte, Shruti Palaskar, Lucas Ventura, Deepti Ghadiyaram, Kenneth DeHaan, Florian Metze, Jordi Torres, and Xavier~Giro i~Nieto.
\newblock How2sign: A large-scale multimodal dataset for continuous american sign language, 2021.

\bibitem[Fang et~al.(2017)Fang, Co, and Zhang]{Fang_2017}
Biyi Fang, Jillian Co, and Mi~Zhang.
\newblock Deepasl: Enabling ubiquitous and non-intrusive word and sentence-level sign language translation.
\newblock In \emph{Proceedings of the 15th ACM Conference on Embedded Network Sensor Systems}, SenSys ’17. ACM, November 2017.
\newblock \doi{10.1145/3131672.3131693}.
\newblock URL \url{http://dx.doi.org/10.1145/3131672.3131693}.

\bibitem[Hendrycks and Gimpel(2016)]{Hendrycks2016GaussianEL}
Dan Hendrycks and Kevin Gimpel.
\newblock Gaussian error linear units (gelus).
\newblock \emph{arXiv: Learning}, 2016.
\newblock URL \url{https://api.semanticscholar.org/CorpusID:125617073}.

\bibitem[Li et~al.(2020)Li, Opazo, Yu, and Li]{li2020wordlevel}
Dongxu Li, Cristian~Rodriguez Opazo, Xin Yu, and Hongdong Li.
\newblock Word-level deep sign language recognition from video: A new large-scale dataset and methods comparison, 2020.

\bibitem[Loshchilov and Hutter(2017)]{Loshchilov2017DecoupledWD}
Ilya Loshchilov and Frank Hutter.
\newblock Decoupled weight decay regularization.
\newblock In \emph{International Conference on Learning Representations}, 2017.
\newblock URL \url{https://api.semanticscholar.org/CorpusID:53592270}.

\bibitem[Martinez et~al.(2002)Martinez, Wilbur, Shay, and Kak]{1166987}
A.M. Martinez, R.B. Wilbur, R.~Shay, and A.C. Kak.
\newblock Purdue rvl-slll asl database for automatic recognition of american sign language.
\newblock In \emph{Proceedings. Fourth IEEE International Conference on Multimodal Interfaces}, pages 167--172, 2002.
\newblock \doi{10.1109/ICMI.2002.1166987}.

\bibitem[Müller et~al.(2020)Müller, Kornblith, and Hinton]{müller2020does}
Rafael Müller, Simon Kornblith, and Geoffrey Hinton.
\newblock When does label smoothing help?, 2020.

\bibitem[Papineni et~al.(2002)Papineni, Roukos, Ward, and Zhu]{papineni-etal-2002-bleu}
Kishore Papineni, Salim Roukos, Todd Ward, and Wei-Jing Zhu.
\newblock {B}leu: a method for automatic evaluation of machine translation.
\newblock In Pierre Isabelle, Eugene Charniak, and Dekang Lin, editors, \emph{Proceedings of the 40th Annual Meeting of the Association for Computational Linguistics}, pages 311--318, Philadelphia, Pennsylvania, USA, July 2002. Association for Computational Linguistics.
\newblock \doi{10.3115/1073083.1073135}.
\newblock URL \url{https://aclanthology.org/P02-1040}.

\bibitem[Shi et~al.(2022)Shi, Brentari, Shakhnarovich, and Livescu]{shi2022opendomain}
Bowen Shi, Diane Brentari, Greg Shakhnarovich, and Karen Livescu.
\newblock Open-domain sign language translation learned from online video, 2022.

\bibitem[Tarrés et~al.(2023)Tarrés, Gállego, Duarte, Torres, and i~Nieto]{slt-how2sign-wicv2023}
Laia Tarrés, Gerard~I. Gállego, Amanda Duarte, Jordi Torres, and Xavier~Giró i~Nieto.
\newblock Sign language translation from instructional videos.
\newblock In \emph{Proceedings of the IEEE/CVF Conference on Computer Vision and Pattern Recognition (CVPR) :Workshops}, 2023.

\bibitem[Yin and Read(2020)]{yin2020better}
Kayo Yin and Jesse Read.
\newblock Better sign language translation with stmc-transformer, 2020.

\bibitem[Zahedi et~al.(2005)Zahedi, Keysers, Deselaers, and Ney]{zahedi2005combination}
Morteza Zahedi, Daniel Keysers, Thomas Deselaers, and Hermann Ney.
\newblock Combination of tangent distance and an image distortion model for appearance-based sign language recognition.
\newblock In \emph{Pattern Recognition: 27th DAGM Symposium, Vienna, Austria, August 31-September 2, 2005. Proceedings 27}, pages 401--408. Springer, 2005.

\end{thebibliography}






\end{document}